# Semi-autonomous Robotic Disassembly Enhanced by Mixed Reality

Alireza Rastegarpanah[1]*

School of Metallurgy and Materials, University of Birmingham, Birmingham B15 2TT UK, a.rastegarpanah@bham.ac.uk

Cesar Alan Contreras*

School of Metallurgy and Materials, University of Birmingham, Birmingham B15 2TT UK, cac214@bham.ac.uk

Rustam Stolkin[1]

School of Metallurgy and Materials, University of Birmingham, Birmingham B15 2TT UK, r.stolkin@bham.ac.uk

In this study, we introduce "SARDiM," a modular semi-autonomous platform enhanced with mixed reality for industrial disassembly tasks. Through a case study focused on EV battery disassembly, SARDiM integrates Mixed Reality, object segmentation, teleoperation, force feedback, and variable autonomy. Utilising the ROS, Unity, and MATLAB platforms, alongside a joint impedance controller, SARDiM facilitates teleoperated disassembly. The approach combines FastSAM for real-time object segmentation, generating data which is subsequently processed through a cluster analysis algorithm to determine the centroid and orientation of the components, categorizing them by size and disassembly priority. This data guides the MoveIt platform in trajectory planning for the Franka Robot arm. SARDiM provides the capability to switch between two teleoperation modes: manual and semi-autonomous with variable autonomy. Each was evaluated using four different Interface Methods (IM): direct view, monitor feed, mixed reality with monitor feed, and point cloud mixed reality. Evaluations across the eight IMs demonstrated a 40.61% decrease in joint limit violations using Mode 2. Moreover, Mode 2-IM4 outperformed Mode 1-IM1 by achieving a 2.33%-time reduction while considerably increasing safety, making it optimal for operating in hazardous environments at a safe distance, with the same ease of use as teleoperation with a direct view of the environment.

**ACM Reference Format:**
---

---

[1] The Faraday Institution, Quad One, Harwell Science and Innovation Campus, Didcot, OX11 0RA, UK.

* Alireza Rastegarpanah and Cesar Alan Contreras are identified as joint first authors.

Semi-autonomous Robotic Disassembly Enhanced by Mixed Reality

# 1 INTRODUCTION

In the industrial sector, the demand for disassembly techniques is growing, fuelled by environmental considerations and regulations. Teleoperated disassembly is relevant for tasks that require high levels of dexterity. This study employs the case of Lithium-Ion Batteries (LIBs) as a specific application to demonstrate the capabilities of the proposed teleoperation framework. LIBs have become crucial in the advancement of alternative energy solutions, most notably in energy storage systems and in electric vehicles (EVs) [1]. The disassembly and dismantling of LIBs involve many challenges stemming from the diversity in battery models, sizes, shapes, and conditions, a variety of actions compound challenges a robot must execute, such as cutting, pulling, unbolting, and sorting. Factors like safety, repeatability, efficiency, and adaptability further increase the complexity of the disassembly process, making it labour-intensive, prone to errors, costly, and potentially hazardous, especially when dealing with damaged batteries [2]. The disassembly of EV batteries commences at the pack level, progresses to the module level, and ends at the cell level.

In most cases, EV battery disassembly remains a manual process carried out by technicians due to the low costs and high levels of manipulation and dexterity required. This manual approach presents risks including fire hazards, explosions, electric shocks, and toxic gas releases. Efforts are being made to overcome the drawbacks of traditional disassembly methods and to improve precision, which is particularly important for preserving valuable components, while also maintaining costs. Teleoperated robotic disassembly emerges as a suitable solution for tasks requiring adaptability, contributing to enhanced safety and time efficiency [2]. Disassembly also faces challenges related to visual perception, automation, and precision [3]. Haptic devices and force feedback systems improve the operator's situational awareness [4]. Virtual Reality (VR) and Mixed Reality (MR) have added value to disassembly scenarios by providing immersive visualisations that support remote operation [5]. These immersive technologies supplement teleoperation methods and are further enhanced by simulations and digital twins. Computer vision techniques, combined with algorithms such as YOLO, have demonstrated their effectiveness in identifying regions of interest, thus aiding in robot planning [6].

In the study, SARDiM, a semi-autonomous robotic disassembly framework, is presented to extend safety in telerobotics, facilitate operators` control in hazardous situations and increase the ease of use of teleoperation. The framework integrates mixed reality, object segmentation via FastSAM [7], and trajectory planning using MoveIt [8]. A Master-Slave joint-joint telerobotics paradigm, utilising similar robots, allows for remote control and force-torque data transfer between the robots. To the best of our knowledge, no other study has proposed a comparable framework integrating a coupled Master-slave haptic interface with mixed reality, capable of executing tasks across multiple autonomy levels. Similarly, there are no platforms known to us that integrate a real-time segmentation method with an adaptive task planner, enabling the execution of complex tasks in unstructured environments. Our research specifically focuses on conducting a comparative analysis between various interface methods and autonomy modes implemented within the SARDiM framework. The comparisons aim to identify how different ways of observing a scene and levels of autonomy impact the efficiency, safety, and user-friendliness of Master-Slave teleoperated tasks, particularly those involving hazardous environments.

# 2 RELATED WORKS

Recently progress in utilising collaborative robots in disassembly has been made with developments in several areas significant to teleoperation; vision, task planning, variable autonomy; drawn by the high demand in disassembly and other manufacturing tasks, which require a safer, more efficient, and versatile process.

## 2.1 Robot Vision

Accurate object identification and proper segmentation are key factors in ensuring successful task execution. A method utilised by Wu et al. [9], employing a simulation-to-reality point cloud 3d object localisation approach to localize screws for disassembly. While it was effective in simulation, the method exhibited a 12.95% decline in detection accuracy when applied to the real-world scenario, also being limited by the dataset size required for training. In another study by Wong et al. [10], the Mask Region-Convolutional Neural Network (Mask R-CNN) and Conditional Random Fields (CRFs) were used for key point and object detection, achieving sub-centimetre error but requiring training on both background and specific items. Yildirim et al. [11] also proposed an object detection algorithm based on an artificial neural network using a custom model optimized for refuelling applications. This was applied to a 6 DoF robot arm, which achieved a 96.6% accuracy rate in the refuelling task. This method is also applicable to other object localisation tasks but is constrained by the need for extensive labelling of specific objects for detection, hindering the adaptability in tasks characterised by uncertainty, like battery pack disassembly with variable models. To overcome this limitation, SARDiM integrates the FastSAM model, which offers precise object segmentation and localisation, and has the capability of segmenting objects without the need for extra training, allowing users to obtain the location of the centre of mass of any object of interest in the scene.

## 2.2 Task Planning

In robotic disassembly, effective planning of end-effector trajectories is crucial and often aided by vision algorithms. Chouex et al. [12] employed vision algorithms in combination with hierarchical task planning to disassemble a battery pack in steps; their approach yielded an average task completion time of 34 seconds with errors of less than 5mm but lacked flexibility in both task sequencing and trajectory adjustments, as their system followed a pre-defined order and path. Lee et al. [13] addressed this adaptability constraint by incorporating a human in the loop, still utilising task planning, but allocating different tasks to the human and the robot. While a human in the loop allows for humans to perform their tasks with some extra adaptability, it compromises safety, especially in hazardous tasks like battery disassembly. SARDiM addresses these limitations by enabling the user to resequence tasks based on a hierarchy and to assume control of the robot via teleoperation during any disassembly tasks. This enhances adaptability to varying models and scenarios.

## 2.3 Teleoperation and Variable Autonomy

Teleoperation in robotic systems offers significant benefits, enabling operators to handle hazardous tasks remotely, thereby minimising risk to human life. Levels of autonomy in teleoperated systems can vary. They can be fully manual, semi-autonomous—where the robot performs certain tasks autonomously while the operator handles more delicate tasks—or largely autonomous, requiring operator intervention only in emergency situations.

Bustamante et al. [14] used a Master-Slave teleoperated robotic system to control a robot with a joystick. Assisted by an Action Template planning framework with various degrees of autonomy, they successfully grabbed objects in a scene with a 95% success rate. While the levels of autonomy allowed the system to adapt to variations of its experiment, it lacked haptic feedback and did not account for joint space limits.

In the literature, multiple teleoperation controllers have been explored. Some of the most used controllers include Cartesian and joint space control. Kebria et al. [15] used these to control a UR5 in simulation, directly comparing both controls. Their study showed that Cartesian control is more prone to singularities in master-slave teleoperation with different DOFs, making joint space control a better alternative. Singh et al. [16] employed identical twin Franka Emika Panda Robot arms equipped with integrated force-torque sensors for master-slave teleoperation. They used a joint torque coupling controller to solve the force feedback problem. However, their tests were limited to easily manipulable objects and did not include more challenging geometries to verify capabilities. Additionally, their controller did not allow varying force limits, which can be inconvenient for tasks that require delicate use of force. SARDiM includes haptic feedback and joint impedance control with identical twin master-slave robots. It also offers variable control of autonomy levels, which has proven useful in an industrial context [17]. Furthermore, SARDiM includes the ability to vary maximum gripping force and maximum movement speed.

**2.4 Mixed Reality Systems**

Mixed Reality Systems can incorporate environmental perception, touch, vision, and sound, immersing an operator further into a scene and thereby enhancing the operator's overall perception of a task.

The study by White et al. [18] showed that operators preferred to use graphical user interfaces (GUIs) for teleoperation robot tasks. This preference reduced frustration in tasks with higher levels of difficulty. Their system's error recovery capability scored 4 out of 5 on the Likert scale for its usefulness within the tested participants. However, this experiment was limited to recovery from a single manipulation task. To account for this SARDiM includes a GUI with buttons allowing error recovery from any robot state. In another study, Sun et al. [19] demonstrated how MR and VR can enhance telepresence systems and reduce operational workload, cutting task times by up to 30%. They used a digital twin to emulate a real robot's movements, limited by the twin not providing the user with any force feedback. De Pace et al. [20] used a similar MR method, generating point clouds in real-time and creating a twin of the robot and the scene but not allowing the robot to be operated with direct user interaction. SARDiM overcomes these limitations by overlaying the digital twin directly on top of the real robot, also enabling the user to feel force feedback and control the robot with joint impedance control. It also introduces the ability for the digital twin to perform the task before the real robot, confirming successful path planning. Although SARDiM is a robust framework, it has a limitation: it requires significant computational power to process its capabilities. Because of this, the processes of the system are distributed among three computers. The processes running within each computer are described in *Figure 1*.

**3 METHODOLOGY**

SARDiM addresses common shortcomings in related works by eliminating the need for direct in-scene human intervention in manufacturing tasks and offering better adaptability through its control methods, force feedback, and general object segmentation. The framework enhances immersion using a mixed reality environment and serves as a comprehensive solution for teleoperation in terms of safety, efficiency, adaptability, and overall user experience. It also paves the way for broader applications in various teleoperated settings, especially those characterised by high levels of uncertainty due to diversity in product designs and unknown conditions. While SARDiM is capable of handling various tasks, this case study focuses on the disassembly of a Nissan Leaf EV battery, from stack level to module level.

**3.1 Control Strategy Setup**

A strategy involving the operation of two identical Franka Emika Panda Cobots in a master-slave configuration is employed, doing a 1:1 mapping of both joint position and torque with a joint impedance control. The equations for this control are defined as follows:

Semi-autonomous Robotic Disassembly Enhanced by Mixed Reality

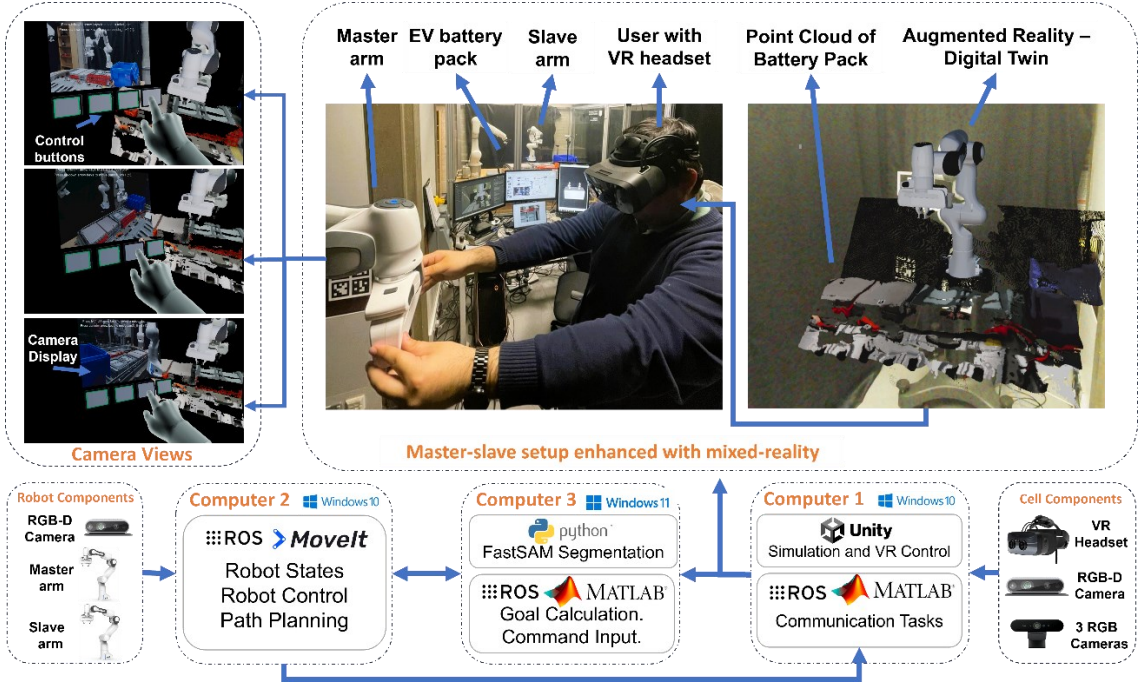

Figure 1: Outline of SARDiM framework: Computer 1 serves as the Mixed Reality and Simulation machine; Computer 2 serves as the Robot Control Machine; and Computer 3 serves as the object detection and goal calculation machine.

$$\tau_f = -K_p e_q - K_d \dot{e}_f + C(q_f) + g(q_f)$$

where $e_q = q_f - q_l$ denotes the joint space error. Here, $K_p$ and $K_d$ are the proportional and derivative gains respectively, while $C(q_f)$ and $g(q_f)$ represent the Coriolis and gravitational components. The closed-loop dynamics in the joint space for the slave arm, that ensure stability and responsiveness, are described by:

$$M \ddot{q}_f + K_d \dot{e}_q + K_p e_q = \tau_{ext}$$

This dynamic equation accentuates the relationship between the mass matrix $M$ of the robot, joint acceleration $\ddot{q}_f$ the error dynamics, and the external torques applied to the system. Furthermore, for the master control, the torques are computed as:

$$\tau_l = \tau_{ext} - K_{dl} \dot{q}_l$$

In this equation, $\tau_{ext}$ represents the external torque estimated to be applied on the slave arm, and $K_{dl}$ is an added damping term for the master's response, useful for managing the feedback effects arising from the bilateral master-slave coupling due to the low stiffness of the human operator. Tuning of the control gains $K_p$ and $K_d$ was executed through a series of preliminary experiments, which minimised the impact of uncertainties in the model parameters $C$ and $g$. For the joint control strategy, the gains were scaled in accordance with the torque capabilities of each joint.

### 3.2 Experimental setup

*Figure 2* shows the workcell setup, which includes three RGB cameras for partial scene views at different angles, a Franka Robot arm as the slave robot, and two Realsense 435i (RGBD) cameras. One Realsense camera is stationed on a stand to scan the workspace, while the other is mounted on the slave robot to perform eye-to-hand calculations for object positioning. SARDiM hardware configuration is made of i) Computer 1, which operates on Windows 10 with an RTX 3080 Supreme graphics card, an Intel i7 processor, and 32 GB of RAM, ii) Computer 2 uses Ubuntu and features a GTX 1080 Ti graphics card, an Intel i7 processor, and 32 GB of RAM, and iii) Computer 3 runs on Windows 11 and is equipped with an RTX 4080 graphics card, an Intel i7 processor, and 64 GB of RAM. Specific processes, such as real-time segmentation, are allocated to more powerful machines to ensure the best performance.

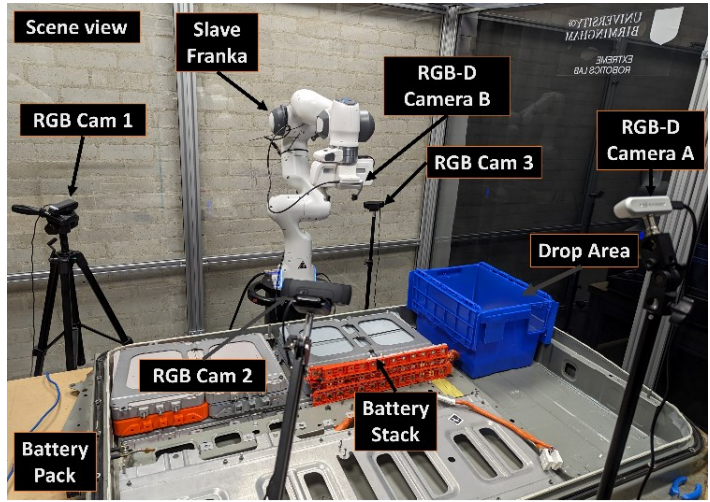

Figure 2: Experimental setup featuring a Nissan Leaf Battery pack as the test subject and a Franka Robot Arm functioning as the slave robot for teleoperation tasks.

### 3.3 Disassembly Tasks and Operation Modes

The case study focuses on the disassembly of an EV battery from the stack to the module level. The task involved the removal and sorting of 8 bolts, a metallic cover case, a plastic bus bar, and battery modules into a blue disposal area adjacent to the stack. SARDiM's performance in the disassembly process is evaluated through key metrics such as average task completion time and occurrences of operator-induced joint limit violations. These metrics are recorded for each specific task. The battery disassembly experiment involves two distinct modes of operation within SARDiM, each featuring four different interface methods (IM). The modes of operation are: **Mode 1: Full Manual Control** - In this mode, the user manually operates the master robot. The slave robot mirrors these movements using a joint impedance controller, **Mode 2: Semi-Autonomous with Variable Autonomy** - The master robot autonomously navigates to a user-defined offset from the targeted object's centre of mass $CM$ that is located within the generated $LST$, calculated by *Algorithm 1* with the automatic FastSAM segmentation. Upon nearing the target, the user is given manual control of the robot to complete the task.

The four interface methods (IM) or methods of scene view are defined as follows: ***IM1 - Direct View:*** The user observes the setup directly, including both master and slave robots. ***IM2 - Monitor Feed:*** The user's view is limited to monitors that display feeds from three cameras in the workcell. ***IM3 - Mixed Reality with Monitor Feed:*** The user employs a VR headset, which receives feed from three camera views of the scene (*Figure 1*). ***IM4 - Point Cloud Mixed Reality:*** Camera feeds are overlaid on the scene and in the hand of the operator. A digital twin is overlaid on the master robot, and a generated point cloud is overlaid in front of the robot. This configuration provides a mixed reality representation of the workcell (*Figure 1*).

---

**ALGORITHM 1: Cluster Analysis**

**Input:** Matrix $S$ containing Segmented Object Matrices $Obj$ and Priority Levels $pL$
**Output:** Metrics List $LST$
**for** each Segmented Object Matrix $Obj$ and corresponding $pL$ in Matrix $S$ **do**
    Cluster of Coordinates: $CRD \leftarrow \{(i,j) \mid Obj_{i,j} = 1\}$
    Get Cluster Area: $A \leftarrow \sum_{(i,j) \in CRD} 1$
    Center of Mass: $CM \leftarrow \frac{1}{A} \sum_{(i,j) \in CRD} (i,j)$
    Center Coordinates: $CCRD \leftarrow \{(i,j) - CM \;\forall\; (i,j) \in CRD\}$
    Covariance Matrix:} $CV \leftarrow \frac{1}{A} CCRD^T CCRD$
    Solve Characteristic Equation: $\det(CV - \lambda I) = 0$ to find $\lambda$
    Find Eigenvectors: Solve $(CV - \lambda I)v = 0$ for $v$
    Major Axis Eigenvector: $AXIS \leftarrow v_{argmax(\lambda)}$
    Rotation Angle $\theta \leftarrow arctan2(AXIS_y, AXIS_x)$
    Update Metrics List: $LST \leftarrow LST \cup \{CM \cup \{pL, \theta, A\}\}$
**End for**
Return $LST$

---

### 3.4 SARDiM Architecture

SARDiM's architecture involves multiple interconnected processes, executed across three computers as delineated in *Figure 1* and detailed as follows: 1) **Object Segmentation and Center of Mass Calculation:** Using Matlab the computer obtains real-time data from the RGBD scene camera via a Transmission Control Protocol (TCP) server. FastSAM segments data generating Matrix $S$ which contains the object matrices $Obj$ and the user-generated priority levels $pL$, which is then processed through *Algorithm 1*, to calculate the centres of mass $CM$, orientations $\theta$ and areas $A$ of the segmented objects, returned on the list $LST$. The process in the algorithm also involves obtaining the clusters of the segmented coordinates ($CRD$), centring them ($CCRD$), getting the covariance matrix ($CV$), the characteristic polinomial ($\lambda$), and the major axis ($AXIS$). 2) **Hierarchical Planning Structure**: Using the segmented data, which includes each object's area, centroid, orientation, and priority, it prepares messages—object descriptors—for trajectory planning via MoveIt. Within the same priority level, planning is biased towards larger objects, ranking them higher in the hierarchy. 3) **Robot Arm Trajectory Planning**: Utilizes the MoveIt joint controller to generate trajectories from a start point to a target, incorporating offsets to the target previously set by the operator. Trajectories can be visualised in the digital twin or executed directly in both master and slave robots. 4) **Mirrored Joint Control**: Provides joint impedance control, mirroring operator-controlled master arm joint movements to the follower robot arm, augmented with force feedback. Mode activates post-execution of a planned trajectory if Mode 2 is enabled. If Mode 1 is active, joint impedance control remains continuously on. 5) **Variable Autonomy**: Control mode commands are received and allow the user to change between manual and semi-autonomous mode. This functionality is initiated through a user interface button in MR and can be run anytime. 6) **Mixed Reality**: Connects to various RGB cameras for MR visual feedback and generates a 3D point cloud from the RGBD camera data positioned in the scene. Synchronises and positions MR elements by utilizing markers affixed to the master robot and the scene, serving as a reference point for real-world localisation. 7) **User Interaction**: Activated by holding the left arm in front of the headset, the interface features four buttons: Mode control, trajectory execution, and two for camera controls (*Figure 1*).

## 4 EXPERIMENTAL RESULTS

The performance of SARDiM in disassembling the battery stack to the module level was assessed across eight trials for each interface mode (IM). Notably, comparisons with other methods were not conducted due to SARDiM's unique features. *Table 1* reports key metrics including average disassembly time, average count of successfully sorted bolts, and average instances of operator-induced joint limit violations, which occur when an operator loses track of the robot arm's joint positions, inadvertently rotating the joints beyond their operational limits, thereby triggering the system's safety mechanisms, and causing a halt in movement. Metrics for sorting additional components (metallic case cover, bus bar, and battery cells) are excluded from the table; these tasks achieved a 100% success rate across all trials. *Table 2* documents the time duration (in mm:ss format) required for each task, on the different IMs. *Table 3* presents a comparative analysis between manual control and variable autonomy modes. It details the percentage increase (+) or decrease (-) in the average time taken, bolts sorted, and operator-induced joint limit violations for corresponding modes.

Table 1: Task Performance (mm:ss)

| IM | Average Time | Average Sorted Bolts | Average Joint Limit Violations | Average Time | Average Sorted Bolts | Average Joint Limit Violations |
|---|---|---|---|---|---|---|
| | Mode 1 | | | Mode 2 | | |
| **IM1** | 09:58 | 6.71 | 2.14 | 10:15 | 6.86 | 1.29 |
| **IM2** | 11:41 | 6.29 | 1.86 | 11:42 | 6.43 | 1.29 |
| **IM3** | 11:41 | 6.43 | 1.71 | 11:33 | 6.43 | 1 |
| **IM4** | 09:49 | 4.86 | 1.71 | 09:44 | 5 | 0.86 |

Table 2: Tasks Average Times (mm:ss)

| IM | Bolts | Busbar | Cover Case | Battery Cells | Bolts | Busbar | Cover Case | Battery Cells |
|---|---|---|---|---|---|---|---|---|
| | Mode 1 (Fully manual) | | | | Mode 2 (Semi-Auto – Variable Autonomy) | | | |
| **IM1** | 05:33 | 00:38 | 01:08 | 02:38 | 05:38 | 00:35 | 01:05 | 02:55 |
| **IM2** | 07:10 | 00:40 | 01:05 | 02:45 | 06:58 | 00:32 | 01:07 | 02:45 |
| **IM3** | 07:01 | 00:35 | 01:05 | 03:00 | 07:06 | 00:33 | 01:05 | 02:48 |
| **IM4** | 05:23 | 00:33 | 01:03 | 02:48 | 05:25 | 00:39 | 01:02 | 02:36 |

Table 3: Comparison between IMs on the different modes

| Method | Average Time | Average Sorted Bolts | Average Joint Limit Violations |
|---|---|---|---|
| IM1 | +2.85% | +2.13% | -40.00% |
| IM2 | +0.14% | +2.27% | -30.77% |
| IM3 | -1.20% | +0.00% | -41.67% |
| IM4 | -0.76% | +2.94% | -50.00% |
| Average Change | +0.26% | +1.84% | -40.61 |

## 5 DISCUSSION

This section discusses the performance and operational challenges encountered during the evaluation of the SARDiM across the different modes and interface methods. Although Mode2-IM4 demonstrated advantages in reducing joint limit violations and improving overall task time by 2.33% compared to Mode1-IM1. User feedback attributed bolt sorting inefficiency to the low-resolution point cloud in the mixed MR environment, which compromised the visualization of fine details. Enhanced point cloud resolution could position Mode2-IM4 as the optimal choice for efficient, reliable teleoperated disassembly. Broadly, semi-autonomous operation across the various interface methods (IMs) resulted in enhanced operational efficiency, primarily evidenced by fewer joint limit violations, safety is considerably increased with the reduction of joint limit violations, allowing for a smoother flow when performing a task in a hazardous environment, preventing unwanted consequences.

For bolt removal, the challenge came from the small size of components. Operators faced difficulty in positioning, corroborated by extended task completion times in *Table 2*. For monitor-only IMs, the times averaged 7m 4s, while both direct view and point cloud IMs averaged 5m 30s. This reduction in time for this task is attributed to the better environmental perception these IMs provided to the operator. Removal of the busbar presented no major difficulties, being the fastest completed task, with an average of 35.8 seconds. For the metal cover case, its weight and associated force feedback presented transport challenges, but the task was ultimately successful. In the case of battery modules, operators needed more dexterity than with the other objects, having to grab them from their sides, influencing the completion time for this task, although across the different interfaces, no major time differences were observed for this task. IMs under Mode 2 significantly reduced joint limit violations, with an average decrease of 40.61% as shown in *Table 3*. This highlights the role of trajectory planning in ensuring proper joint positioning and smoother operations. Although time efficiency showed no significant improvement, the marked reduction in joint limit violations indicates enhanced operational reliability while performing a task that should not be stopped. The visual feedback in Mode2-IM4 also contributed to this improvement, having the lowest number of Joint limit violations, at 0.86. In contrast, Mode 1 posed challenges related to monitoring individual joint angles, whereas Mode 2 reset joint positions between manipulation of components.

## 6 CONCLUSION

This paper presented the SARDiM framework, contributing to the field of teleoperated disassembly, aimed at enhancing task accuracy and operator situational awareness through the integration of force feedback, mixed reality and point clouds. Applied to the disassembly of Lithium-Ion Batteries (LIBs), SARDiM demonstrated its efficacy across two operational modes—Manual and Semi-Autonomous—and four Interface Methods (IMs). Our findings indicate a modest but impactful 2.33% speed performance improvement in Mode 2 - IM4 (mixed reality with point cloud), compared to the baseline of Mode 1 - IM1 (direct view). More significantly, the implementation of semi-autonomy resulted in an average error reduction of 40.61%, showing the safety benefits of our semi-autonomous approach. While the speed increase is not dramatic, the notable reduction in joint errors represents a significant advancement in operational safety and efficiency. This allows operators to work across various environments and distances without compromising the performance of the baseline mode, thanks to the enhanced safety from reduced joint limit violations and improved positioning accuracy provided by object segmentation. Although our initial case study was on LIBs disassembly, the SARDiM framework's versatility suggests it could be beneficial in a wide range of settings, from minimizing radiation exposure in nuclear environments to providing precise control in delicate medical procedures and enhancing safety in hazardous industrial operations. Future efforts will enhance semi-autonomous modes, develop full autonomy for predictable tasks, and extend the framework to additional case studies, to further prove its effectiveness.

**ACKNOWLEDGMENTS**

This work was supported in part by the **UK Research and Innovation (UKRI) project "Reuse and Recycling of**

Semi-autonomous Robotic Disassembly Enhanced by Mixed Reality

**Lithium-Ion Batteries" (RELiB)** under RELiB2 Grant FIRG005 and RELiB3 Grant FIRG057 and in part by the REBELION project under Grant 101104241.

## REFERENCES

[1] P. U. Nzereogu, A. D. Omah, F. I. Ezema, E. I. Iwuoha, and A. C. Nwanya. 2022. Anode materials for lithium-ion batteries: A review. Appl. Surf. Sci. Adv. 9, 6 (June 2022), 100233. DOI: https://doi.org/10.1016/J.APSADV.2022.100233

[2] Jamie Hathaway, Abdelaziz Shaarawy, Cansu Akdeniz, Ali Aflakian, Rustam Stolkin, and Alireza Rastegarpanah. 2023. Towards reuse and recycling of lithium-ion batteries: tele-robotics for disassembly of electric vehicle batteries. Front. Robot. AI 10, 8 (August 2023), 1179296. DOI: https://doi.org/10.3389/FROBT.2023.1179296/BIBTEX

[3] Lin Zhou, Akhil Garg, Jun Zheng, Liang Gao, and Ki-Yong Oh. 2021. Battery pack recycling challenges for the year 2030: Recommended solutions based on intelligent robotics for safe and efficient disassembly, residual energy detection, and secondary utilization. Energy Storage 3, 3 (June 2021), e190. DOI: https://doi.org/10.1002/EST2.190

[4] Bin Zhao, Shu'an Zhang, Zhonghao Wu, Bo Yang, and Kai Xu. 2020. CombX: Design and experimental characterizations of a haptic device for surgical teleoperation. Int. J. Med. Robot. Comput. Assist. Surg. 16, 1 (February 2020), e2042. DOI: https://doi.org/10.1002/RCS.2042

[5] Abdeldjallil Naceri, Dario Mazzanti, Joao Bimbo, Domenico Prattichizzo, Darwin G. Caldwell, Leonardo S. Mattos, and Nikhil Deshpande. 2019. Towards a virtual reality interface for remote robotic teleoperation. In Proceedings of the 19th International Conference on Advanced Robotics (ICAR 2019), December 2019, 284-289. DOI: https://doi.org/10.1109/ICAR46387.2019.8981649

[6] Guan Zhaoxin, Li Han, Zuo Zhijiang, and Pan Libo. 2022. Design a Robot System for Tomato Picking Based on YOLO v5. IFAC-PapersOnLine 55, 3 (January 2022), 166-171. DOI: https://doi.org/10.1016/J.IFACOL.2022.05.029

[7] Xu Zhao, Wenchao Ding, Yongqi An, Yinglong Du, Tao Yu, Min Li, Ming Tang, and Jinqiao Wang. 2023. Fast Segment Anything. Preprint. arXiv:2306.12156v1 [cs.CV]. Available at: https://arxiv.org/abs/2306.12156v1 (June 2023).

[8] David T. Coleman, Ioan A. Sucan, Sachin Chitta, and Nikolaus Correll. 2017. Reducing the Barrier to Entry of Complex Robotic Software: a MoveIt! Case Study. J. Softw. Eng. Robot. 5, 1 (2017), 3-16. DOI: https://doi.org/10.6092/JOSER_2014_05_01_P3.

[9] Chengzhi Wu, Xuelei Bi, Julius Pfrommer, Alexander Cebulla, Simon Mangold, and Jürgen Beyerer. 2023. Sim2real Transfer Learning for Point Cloud Segmentation: An Industrial Application Case on Autonomous Disassembly. Preprint. arXiv:2301.05033v1 [cs.CV]. Available at: https://arxiv.org/abs/2301.05033v1 (January 2023).

[10] Ching Chang Wong, Li Yu Yeh, Chih Cheng Liu, Chi Yi Tsai, and Hisasuki Aoyama. 2021. Manipulation Planning for Object Re-Orientation Based on Semantic Segmentation Keypoint Detection. Sensors 21, 7 (March 2021), 2280. DOI: https://doi.org/10.3390/S21072280.

[11] Suleyman Yildirim, Zeeshan A. Rana, and Gibert Tang. 2023. Development of Vision Guided Real-Time Trajectory Planning System for Autonomous Ground Refuelling Operations using Hybrid Dataset. Presented at the American Institute of Aeronautics and Astronautics (January 2023). DOI: https://doi.org/10.2514/6.2023-1148. Available at: https://arc.aiaa.org/doi/10.2514/6.2023-1148

[12] Martin Choux, Eduard Marti Bigorra, and Ilya Tyapin. 2021. Task Planner for Robotic Disassembly of Electric Vehicle Battery Pack. Metals 11, 3 (February 2021), 387. DOI: https://doi.org/10.3390/MET11030387. Available at: https://www.mdpi.com/2075-4701/11/3/387/htm

[13] Meng Lun Lee, Sara Behdad, Xiao Liang, and Minghui Zheng. 2022. Task allocation and planning for product disassembly with human–robot collaboration. Robot. Comput.-Integr. Manuf. 76 (August 2022), 102306. DOI: https://doi.org/10.1016/J.RCIM.2021.102306

[14] Samuel Bustamante, Gabriel Quere, Daniel Leidner, Jorn Vogel, and Freek Stulp. 2022. CATs: Task Planning for Shared Control of Assistive Robots with Variable Autonomy. In Proceedings of the IEEE International Conference on Robotics and Automation (ICRA 2022), 3775-3782. DOI: https://doi.org/10.1109/ICRA46639.2022.9811360

[15] Parham M. Kebria, Abbas Khosravi, Saeid Nahavandi, Abdollah Homaifar, and Mehrdad Saif. 2019. Experimental comparison study on joint and cartesian space control schemes for a teleoperation system under time-varying delay. In Proceedings of the IEEE International Conference on Industrial Technology (ICIT 2019), February 2019, 108-113. DOI: https://doi.org/10.1109/ICIT.2019.8755087

[16] Jayant Singh, Aravinda Ramakrishnan Srinivasan, Gerhard Neumann, and Ayse Kucukyilmaz. 2020. Haptic-guided teleoperation of a 7-DoF collaborative robot arm with an identical twin master. IEEE Trans. Haptics 13, 1 (January 2020), 246-252. DOI: https://doi.org/10.1109/TOH.2020.2971485

[17] Guiyang Xin, Carlo Tiseo, Wouter Wolfslag, Joshua Smith, Oguzhan Cebe, Zhibin Li, Sethu Vijayakumar, and Michael Mistry. 2020. Variable Autonomy of Whole-body Control for Inspection and Intervention in Industrial Environments using Legged Robots. In Proceedings of the IEEE International Conference on Automation Science and Engineering (CASE 2020), August 2020, 1415-1420. DOI: https://doi.org/10.1109/CASE48305.2020.9216813

[18] Samuel S. White, Keion W. Bisland, Michael C. Collins, and Zhi Li. 2020. Design of a high-level teleoperation interface resilient to the effects of unreliable robot autonomy. In Proceedings of the IEEE International Conference on Intelligent Robots and Systems (IROS 2020), October 2020, 11519-11524. DOI: https://doi.org/10.1109/IROS45743.2020.9341322

[19] Da Sun, Andrey Kiselev, Qianfang Liao, Todor Stoyanov, and Amy Loutfi. 2020. A New Mixed-Reality-Based Teleoperation System for Telepresence and Maneuverability Enhancement. IEEE Transactions on Human-Machine Systems 50, 1 (February 2020), 55-67. DOI: https://doi.org/10.1109/THMS.2019.2960676

[20] Francesco De Pace, Gal Gorjup, Huidong Bai, Andrea Sanna, Minas Liarokapis, and Mark Billinghurst. 2020. Assessing the Suitability and Effectiveness of Mixed Reality Interfaces for Accurate Robot Teleoperation. In Proceedings of the ACM Symposium on Virtual Reality Software and Technology (VRST), November 2020. ACM, DOI: https://doi.org/10.1145/3385956.3422092